\documentclass[conference]{IEEEtran}
\IEEEoverridecommandlockouts

\usepackage{cite}
\usepackage{amsmath,amssymb,amsfonts}
\usepackage{algorithmic}
\usepackage{graphicx}
\usepackage{textcomp}
\usepackage{xcolor}
\usepackage{lineno,hyperref}
\usepackage{CJKutf8}
\def\BibTeX{{\rm B\kern-.05em{\sc i\kern-.025em b}\kern-.08em
    T\kern-.1667em\lower.7ex\hbox{E}\kern-.125emX}}
\begin{document}
\begin{CJK}{UTF8}{gbsn}
\title{RSMLP: A light Sampled MLP Structure \\ for Incomplete Utterance Rewrite
\thanks{This paper is supported by the National Natural Science Foundation of China under Grant 62473138,  the Project of Natural Science Foundation Youth Enhancement Program of Guangdong Province under Grant 2024A1515030184,  the Project of Guangzhou City Zengcheng District Key Research and Development under Grant 2024ZCKJ01, and the General Project of Natural Science Foundation of Hunan Province under Grant 2022JJ30162. \\ $^{*}$Corresponding author is Weilai Jiang from Hunan University (Jiangweilai@hnu.edu.cn)}
}

\author{
    \IEEEauthorblockN{Lunjun Liu$^{a,b}$, $^{*}$Weilai Jiang$^{a,b}$, Yaonan Wang$^{a,b}$}
    \IEEEauthorblockA{$^a$College of Electrical and Information Engineering, Hunan University, Changsha, P.R.China}
    \IEEEauthorblockA{$^b$Greater Bay Area Institute for Innovation, Hunan University, Guangzhou, P.R.China}
    \IEEEauthorblockA{\{barryyyliu, jiangweilai, yaonan\}@hnu.edu.cn}
}

\maketitle

\begin{abstract}
The Incomplete Utterance Rewriting (IUR) task has garnered significant attention in recent years. Its goal is to reconstruct conversational utterances to better align with the current context, thereby enhancing comprehension. In this paper, we introduce a novel and versatile lightweight method, Rewritten-Sampled MLP (RSMLP). By employing an MLP-based architecture with a carefully designed down-sampling strategy, RSMLP effectively extracts latent semantic information between utterances and makes appropriate edits to restore incomplete utterances. Due to its simple yet efficient structure, our method achieves competitive performance on public IUR datasets and in real-world applications.
\end{abstract}


\section{Introduction}
In recent years, conversation-based tasks have gained increasing attention, such as dialogue response generation \cite{b31}\cite{b32} and dialogue understanding \cite{b33}\cite{b34}. The advent of Large Language Models (LLMs) has shifted the focus from single-turn to multi-turn dialogues. In multi-turn dialogue scenarios, users tend to use incomplete utterances, which often omit or reference entities or concepts from previous dialogue context, a phenomenon known as ellipsis and anaphora. Studies have shown that over 70\% of utterances exhibit these phenomena \cite{b1}, which significantly impacts the accuracy of semantic understanding in dialogue systems.

To address this issue, recent research has introduced the Incomplete Utterance Rewriting (IUR) task \cite{b3}\cite{b4}\cite{b5}. The goal of the IUR task is to rewrite incomplete utterances into new sentences with the same semantics, where the new sentences can be understood without referring to the context. As shown in Table \ref{example1}, $(u_{1}, u_{2}, u_{3})$ form a multi-turn dialogue, where $u_{3}$ is an incomplete utterance that omits "Shenzhen" and uses "this" to refer to "wet". The revised $u_{3}^{*}$ is a complete sentence that can be understood independently. By explicitly rewriting the omitted information into the latest utterance, downstream dialogue models only need to process the final utterance. This significantly alleviates the model's burden during long-term reasoning.

\begin{table}[t]
\label{example1}
  \begin{center}
    \caption{The example of incomplete utterance rewriting
.}
    \begin{tabular}{c|c} 
      \textbf{Turns} & \textbf{Utterance (Translation)} \\
      \hline
      $u_{1}$ & 深圳的气候怎么样\\
       & (How is the climate in Shenzhen) \\
      $u_{2}$ & 十分潮湿\\
      & (It is quite wet) \\
      $u_{3}$ & 为什么会这样\\
      & (Why is this) \\
      $u_{3}^{*}$ & 深圳的气候为什么会十分潮湿\\
      & (Why is the climate in Shenzhen so wet) \\
    \end{tabular}
  \end{center}
  \vspace{-0.5em}
\end{table}

Although significant progress has been made in previous work, balancing the quality of sentence rewriting with autoregressive generation speed remains a challenge for the IUR task. To improve speed, RUN\cite{b6} framed the IUR task as a semantic segmentation problem based on feature mappings constructed from word embeddings. RAU\cite{b7} extracted coreference and ellipsis relationships from the self-attention weight matrices of transformers. Both approaches utilize U-Net, a complex model that significantly impacts rewriting efficiency. To enhance quality, SRL\cite{b10} trained a semantic role labeling model to emphasize the core meaning of each input dialogue, preventing the rewriter from violating key content. They manually annotated SRL information for over 27,000 dialogue turns, a time-consuming and costly process. PAC\cite{b4} constructs a "pick-and-combine" model to extract omitted tokens from the context in order to restore incomplete sentences. RAST\cite{b11} formulated the IUR task as a span prediction problem of deletion and insertion, using reinforcement learning to improve fluency, which heavily depends on the encoder's output.
\begin{figure*}[t]
\centerline{\includegraphics[scale=0.517]{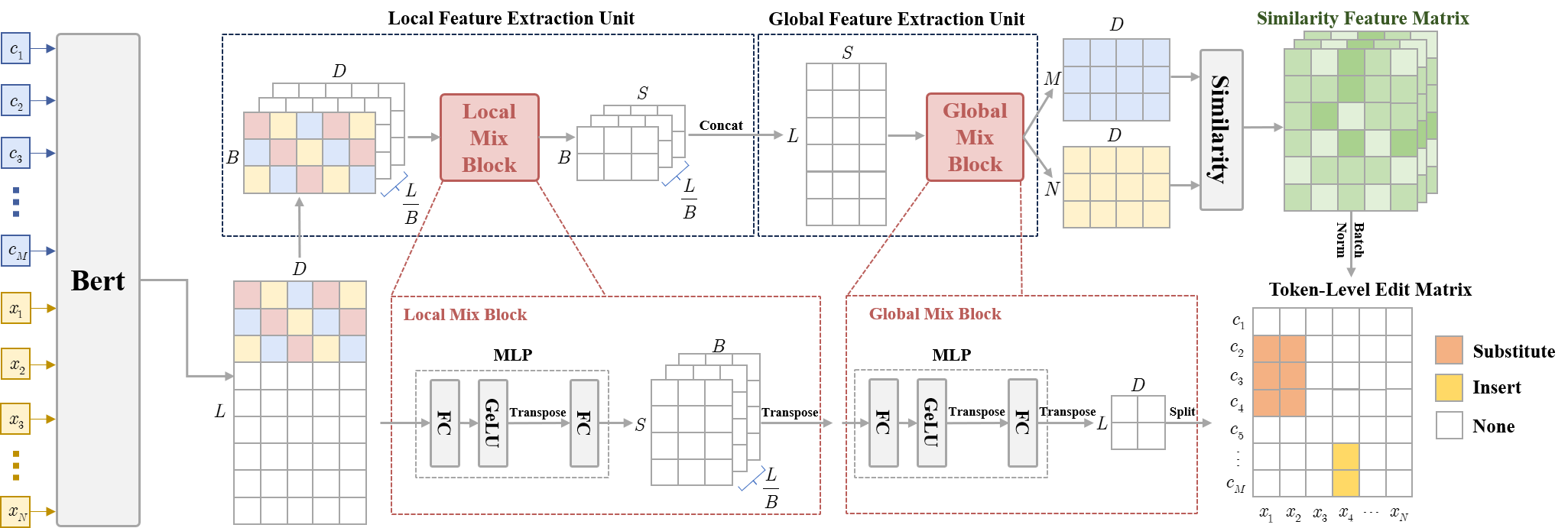}}
\caption{The architecture of RSMLP.}
\label{frame}
\end{figure*}

In this paper, we explore the use of a light neural network architecture for sentence rewriting, specifically utilizing a simple MLP architecture based on a down-sampling strategy. Our approach involves using down-sampling and MLP to sequentially refine local and global semantic information. Subsequently, we perform similarity calculations on the output semantic matrices to obtain similarity feature maps, which are then used to construct a token-level edit matrix. Finally, we edit incomplete utterances based on the predicted edit type tokens to generate the rewritten sentences. Our contributions are summarized as follows:
\begin{itemize}
\item We investigate capturing local information in the IUR task using a down-sampling approach.
\item We propose a model structure, RSMLP, composed solely of MLPs, which is both simple and efficient.
\item Experimental results demonstrate that our method achieves a balance between rewriting quality and inference speed, and performs well in real-world scenarios.
\end{itemize}

\section{Related Works}

\section{Method}

In this section, we will provide a detailed explanation of our method.
\subsection{Problem Definition}
We formally define the utterances that need to be rewritten. For multi-turn dialogue utterances $(u_{1}, u_{2}, \dots, u_{t})$, we concatenate all the context utterances $(u_{1}, u_{2}, \dots, u_{t-1})$ into a word sequence of length $M$, denoted as 
$c=(c_{1},c_{2},\dots,c_{M})$, and use a special [SEP] token to separate utterances from different contexts. The final utterance $u_{t}$ in the dialogue is defined as the incomplete utterance, represented as a word sequence of length $N$, $x = (x_{1}, x_{2}, \dots, x_{N})$.

\subsection{Architecture}
As shown in Figure \ref{frame}, we propose a simple model architecture composed of four components: an Encoder, a Local Feature Extraction Unit, a Global Feature Extraction Unit, and a Similarity Feature Matrix. Since our core concept is an MLP based on a sampling mechanism, we named the model \textbf{R}ewritten-\textbf{S}ampled \textbf{MLP} (RSMLP).

\textbf{Encoder} \quad We use BERT\cite{b12} as our Encoder to extract contextual information between utterances. First, we concatenate the context utterance $c$ with the incomplete utterance $x$, forming a joint token sequence of length $L=M+N$. This sequence is then fed into BERT to produce a word embedding matrix $A \in \mathbb{R}^{L\times D}$.

\textbf{Local Feature Extraction Unit} \quad To capture both local and global information, we divide the core of our framework into two components: the Local Feature Extraction Unit and the Global Feature Extraction Unit. The design of the Local Feature Extraction Unit is primarily inspired by the concept of down-sampling. Since our sentence rewriting task involves editing and replacing parts of the original sentence—similar to classification tasks—down-sampling effectively balances the model's ability to recognize different categories, thereby enhancing its rewriting capabilities. To address the issue of potential information loss caused by naïve down-sampling methods, we introduce continuous sampling, which helps the model capture local information more accurately.

Our sampling method converts each multi-turn dialogue sequence of length $L$ into multiple non-overlapping subsequences of length $B$, resulting in $\frac{L}{B}$ subsequences. The word matrix can then be decomposed as:
\begin{equation}
A = \{A_{1}, A_{2}, \dots, A_{\frac{L}{B}}\}
\end{equation}

where each $A_{i}\in\mathbb{R}^{B\times D}$, $i \in [1, \frac{L}{B}]$ is a submatrix, and $A \in\mathbb{R}^{B\times \frac{L}{B}\times D}$. This segmentation divides the sequence into continuous short fragments, allowing the model to focus more on local semantic information. Additionally, this design improves training and inference efficiency, as the model only processes a small portion of the sequence at a time.

Next, we feed the subsequences into the Local Mix Block, which is the basic building block designed for local information exchange, primarily consisting of MLPs. The Local Mix Block includes two fully connected layers and a non-linear activation function, all applied to 2D matrices. The purpose of the Local Mix Block is to facilitate information exchange along two different dimensions, producing another feature matrix. The resulting matrix can be viewed as the feature extraction of the input matrix. The information exchange along the embedding dimension is referred to as input projection:

\begin{equation}
Z = W_{2}(\sigma(W_{1}A+B_{1}))^{T}+B_{2}
\end{equation}

where $Z=\{Z_{1}, Z_{2}, \dots, Z_{\frac{L}{B}}\} \in \mathbb{R}^{B \times\frac{L}{B}\times S}$ is the output matrix, and $W_{1}$,$B_{1}$,$W_{2}$,$B_{2}$ represent the weights and biases of the first and second fully connected layers in the MLP, respectively. The function $\sigma(\cdot)$ denotes the non-linear activation function, with GeLU\cite{b14} being used in our method. It should be noted that $S$ is smaller than $D$, creating a bottleneck structure.

\textbf{Global Feature Extraction Unit} \quad After refining the local information, we designed the Global Feature Extraction Unit to extract global information. We concatenate the matrices $Z_{i\in[1,\frac{L}{B}]}$ to obtain $Z \in \mathbb{R}^{L\times S}$ Next, we feed $Z$ into the Global Mix Block, which facilitates the exchange of global information across two dimensions:

\begin{equation}
Z^{*} = W_{4}(\sigma(W_{3}Z+B_{3}))^{T}+B_{4}
\end{equation}

Where $Z^{*} \in \mathbb{R}^{L\times D}$ is the output matrix, and $W_{3}$,$B_{3}$,$W_{4}$,$B_{4}$ represent the weights and biases of the first and second fully connected layers in the MLP, respectively. The function $\sigma(\cdot)$ denotes the non-linear activation function. Additionally, we apply padding and replication mechanisms\cite{b29}\cite{b30} to ensure consistency in sequence length.

\textbf{Similarity Feature Matrix} \quad To further capture the correlation between words, we used several similarity functions to encode the relationships. Specifically, for each word embedding $\textbf{E}_{x_{N}}$  in the incomplete utterance and each word embedding $\textbf{E}_{c_{M}}$ in the context utterance, we modeled their relationship using three similarity matrices: dot product similarity, cosine similarity, and bilinear similarity. These matrices together form the Similarity Feature Matrix $\textbf{S}(c_{M}, x_{N})$, as shown below:

\begin{align}
\textbf{S}(c_{M},x_{N}) = [\textbf{E}_{x_{N}} \cdot \textbf{E}_{c_{M}};cos(\textbf{E}_{x_{N}},\textbf{E}_{c_{M}});
\\ bilinear(\textbf{E}_{x_{N}}, \textbf{E}_{c_{M}})]
\end{align}

These similarity feature matrices model the correlation between words from different perspectives.

Finally, we apply BatchNorm\cite{b17} to reduce the dimensionality of matrix $\textbf{S}$. At this stage, each feature vector is mapped to one of three label types: $Substitute$, $Insert$, or $None$, resulting in the generation of a Token-level Edit Matrix.

\subsection{Incomplete Utterance Edit}
Since the existing dataset only contains rewritten sentences, we need a process to automatically derive the Token-level Edit Matrix and use these examples for training. We first identify the Longest Common Subsequence (LCS) between the incomplete and rewritten utterances. Then, we enumerate the alignment among the incomplete utterance, rewritten utterance, and the LCS. For words in the rewritten sentence that are not in the LCS, they are labeled as [ADD]. Conversely, for words in the incomplete sentence but not in the LCS, they are labeled as [DEL]. Consecutive words with the same label are merged into a span. By comparing spans, any [DEL] span in the rewritten sentence that corresponds to an [ADD] span in the same context is marked as $Substitute$. Otherwise, the added span is considered an $Insert$, while words that do not undergo any changes are labeled as $None$.

\section{Experiments}

\subsection{Experimental Setup}
\textbf{Datasets} \quad We conducted experiments on IUR benchmarks from three different domains and languages: Restoration - 200k\cite{b4}, REWRITE \cite{b1}, and CANARD \cite{b18}. The datasets were split into training, evaluation, and testing sets with the following proportions: 80\%/10\%/10\% for Restoration - 200k, 90\%/10\%/- for REWRITE, and 80\%/10\%/10\% for CANARD. These datasets consist of multi-turn dialogue contexts, incomplete sentences to be rewritten, and examples of correct rewrites.

\textbf{Baselines} \quad We compared the performance of RSMLP with the following methods: transformer-based pointer generator (T-Ptr-Gen)\cite{b25}, Seq2Seq model L-Gen\cite{b26}, the hybrid pointer generator (L-Ptr-Gen)\cite{b25}, L-Ptr-$\lambda$/T-Ptr-$\lambda$ \cite{b1}, PAC\cite{b4}, CSRL\cite{b9}, SARG\cite{b20}, RAST\cite{b11} and RUN (BERT)\cite{b6}. For details on the benchmarks, please refer to the respective papers.

\textbf{Evaluation} \quad Following previous practices, we used BLEU\cite{b21}, ROUGE\cite{b22}, Exact Match (EM), and Restoration Score\cite{b4} as automatic evaluation metrics to compare our proposed method with other approaches.

\textbf{Model Setting} \quad We used bert-base-chinese from the HuggingFace community\cite{b23} as our pre-trained BERT model and fine-tuned it as part of the training process.  The model has 12 layers and 12 attention heads. We optimized the model using Adam\cite{b24} with a learning rate of 1e-5 and computed the loss using weighted cross-entropy.

\subsection{Main Results}
\begin{table*}[t!]
  \begin{center}
    \caption{The results of all compared models trained and evaluated on the RESTORATION. }
    \label{e1}
    \resizebox{17.8cm}{!}{\begin{tabular}{cccccccccccccc} 
      \textbf{Model} & $\mathcal{P}_{1}$ & $\mathcal{R}_{1}$ & $\mathcal{F}_{1}$ & $\mathcal{P}_{2}$ & $\mathcal{R}_{2}$ & $\mathcal{F}_{2}$ & $\mathcal{P}_{3}$ & $\mathcal{R}_{3}$ & $\mathcal{F}_{3}$ & $\textbf{B}_{1}$ & $\textbf{B}_{2}$ & $\textbf{R}_{1}$ & $\textbf{R}_{2}$\\
      \hline
      T-Ptr-$\lambda$ & - & - & 51.0 & - & - & 40.4 & - & - & 33.3 & 90.3 & 87.4 & 90.1 & 83.0\\
      L-Gen & 65.5 & 40.8 & 50.3 & 52.2 & 32.6 & 40.1 & 43.6 & 27.0 & 33.4 & 84.9 & 81.7 & 88.8 & 80.3 \\
      L-Ptr-Gen & 66.6 & 40.4 & 50.3 & 54.0 & 33.1 & 41.1 & 45.9 & 28.1 & 34.9 & 84.7 & 81.7 & 89.0 & 80.9\\
      PAC & 70.5 & 58.1 & 63.7 & 55.4 & 45.1 & 49.7 & 45.2 & 36.6 & 40.4 & 89.9 & 86.3 & 91.6 & 82.8 \\
      SARG & - & - & 62.4 & - & - & 52.5 & - & - & 46.3 & 92.2 & 89.6 & 92.1 & 86.0\\
      RAST & - & - & - & - & - & - & - & - & - & 90.4 & 89.6 & 91.2 & 84.3 \\
      RUN (BERT) & 73.2 & \textbf{64.6} & 68.6 & 59.5 & 53.0 & 56.0 & 50.7 & 45.1 & 47.7 & 92.3 & 89.6 & 92.4 & 85.1\\
      \textbf{RSMLP (Ours)} & \textbf{76.4} & 64.0 & \textbf{69.6} & \textbf{62.9} & \textbf{53.1} & \textbf{57.3} & \textbf{54.5} & \textbf{46.1} & \textbf{49.7} & \textbf{93.3} & \textbf{90.2} & \textbf{92.5} & \textbf{86.1} \\
      \hline
    \end{tabular}}
    \\{\footnotesize \textsuperscript{*}Note: All results are taken from the original papers. Dashes: results are not reported in the responding literature.}
  \end{center}
  \vspace{-1.0em}
\end{table*}

Tables \ref{e1} and \ref{e2} present the experimental results on the Restoration-200K and Rewrite datasets, respectively. For the Restoration dataset, our proposed RSMLP model outperforms the previously best-performing model, RUN (BERT), on nearly all metrics. In particular, the metrics $\mathcal{P}_{1}$, $\mathcal{P}_{2}$, and $\mathcal{P}_{3}$ show an average improvement of 3.5 points, while $\mathcal{F}_{1}$, $\mathcal{F}_{2}$, and $\mathcal{F}_{3}$ also exhibit significant gains. This demonstrates the effectiveness of our architecture, as RSMLP successfully captures both local and global information, thereby enhancing the rewriting capability. Furthermore, although the differences are small, RSMLP also surpasses previous models in terms of BLEU and ROUGE scores, supporting the robustness of our model.

For the Rewrite dataset, RSMLP similarly achieves better performance across nearly all metrics. Notably, our method improves the EM score by 1.5 points. This indicates that the rewritten sentences generated by our model perfectly match the reference sentences, showcasing its deep understanding of contextual semantics.

\begin{table}[t]
 \vspace {-1.0em}
  \begin{center}
    \caption{The results of all compared models trained and evaluated on the REWRITE. }
    \label{e2}
    \resizebox{8.5cm}{!}{
    \begin{tabular}{cccccc} 
      \textbf{Model} & $\textbf{EM}$ & $\textbf{B}_{2}$ & $\textbf{B}_{4}$ & $\textbf{R}_{2}$ & $\textbf{R}_{L}$\\
      \hline
      L-Gen & 47.3 & 81.2 & 73.6 & 80.9 & 86.3 \\
      L-Ptr-Gen & 50.5 & 82.9 & 75.4 & 83.8 & 87.8 \\
      L-Ptr-$\lambda$ & 42.3 & 82.9 & 73.8 & 81.1 & 84.1 \\
      T-Ptr-$\lambda$ & 52.6 & 85.6 & 78.1 & 85.0 & 89.0 \\
      T-Ptr-Gen & 53.1 & 84.4 & 77.6 & 85.0 & 89.1  \\
      RUN (BERT) & 66.4 & 91.4 & 86.2 & 90.4 & \textbf{93.5}\\
      \textbf{RSMLP (Ours)} & \textbf{67.9} & \textbf{91.5} & \textbf{86.5} & \textbf{90.7} & 93.4\\ 
      \hline
    \end{tabular}}
    \\{\footnotesize \textsuperscript{*}Note: \textbf{EM} indicates the exact match score and $\textbf{R}_{L}$ is ROUGE score based
on the LCS.}
  \end{center}
    \vspace {-1.0em}
\end{table}

\subsection{Inference Speed}
Table \ref{e3} presents the inference speed results for a single sentence. All models were run on a single NVIDIA 3070 Laptop, implemented using PyTorch. We can observe that compared to the state-of-the-art (SOTA) methods, our RSMLP model achieves the fastest inference speed. Specifically, it is 21 times faster than T-Ptr-Gen (n$\_$Beam=1). Furthermore, compared to the second-fastest model, RUN (BERT), our approach also demonstrates superior speed. This highlights that our lightweight MLP architecture significantly enhances sentence inference speed while maintaining high rewriting quality.

\begin{table}[t]
\vspace{-1.0em}
  \begin{center}
    \caption{The inference speed comparison between ReMod and baselines on CANARD.}
    \label{e3}
    \resizebox{6cm}{!}{
    \begin{tabular}{cc} 
      \textbf{Model} & \textbf{Speedup}\\
      \hline
      T-Ptr-Gen (n$\_$Beam=1) & 1$\times$ \\
      T-Gen (n$\_$Beam=1) & 2$\times$ \\
      L-Gen (n$\_$Beam=1) & 4$\times$ \\
      L-Ptr-Gen (n$\_$Beam=1) & 4$\times$ \\
      SARG (n$\_$Beam=1) & 18$\times$ \\
      RUN (BERT) & 18$\times$ \\
      \textbf{RSMLP (Ours)} & \textbf{21}$\times$ \\ 
      \hline
    \end{tabular}}
    \\{\footnotesize \textsuperscript{*}Note: n$\_$Beam refers to the beam size used in beam search, which is not applicable to the RUN and ReMod models.}
  \end{center}
  \vspace {-1.0em}
\end{table}

\subsection{Ablation Study}
To validate the effectiveness of the Local Feature Extraction Unit (LU) and Global Feature Extraction Unit (GU), we conducted a comprehensive ablation study, as presented in Table \ref{e4}.

As expected, the absence of both feature extraction units resulted in a decrease across all metrics, demonstrating that our framework significantly enhances model performance. Moreover, using only one of the units also yielded suboptimal results due to the lack of understanding of certain aspects of the information. This further confirms that RSMLP achieves outstanding performance only when both local and global semantic information are simultaneously utilized.

\begin{table}[t!]
\vspace{-1.0em}
  \begin{center}
    \caption{The ablation study on REWRITE dataset.}
    \label{e4}
    \resizebox{7.5cm}{!}{
    \begin{tabular}{cccccc} 
      \textbf{Model} & \textbf{EM} & $\mathcal{F}_{2}$ & $\mathcal{P}_{2}$ & $\textbf{B}_{2}$ & $\textbf{R}_{2}$\\
      \hline
      RSMLP & 67.9 & 82.6 & 86.5 & 91.5 & 90.7\\
      \hline
      w/o GU & 66.7 & 81.5 & 85.2 & 90.8 & 90.5\\
      w/o LU & 66.4 & 81.1 & 85.3 & 90.6 & 90.2\\
      w/o both & 65.0 & 80.1 & 82.3 & 90.3 & 89.6\\
      \hline
    \end{tabular}
    }    
    \\{\footnotesize \textsuperscript{*}Note: `w/o' stands for `without'.}
  \end{center}
\vspace {-1.0em}
\end{table}

\begin{table}[t!]
\vspace{-1.0em}
  \begin{center}
    \caption{The real-world Experiment}
    \label{e5}
    \resizebox{8cm}{!}{
    \begin{tabular}{cccc} 
      \hline
      \textbf{Model} & \textbf{ROM} & \textbf{Inference Speed} & \textbf{Accuracy} \\
      \hline
      RSMLP & 368MB & 70ms & 96\% \\
      \hline
    \end{tabular}
    }    
    \\{\footnotesize \textsuperscript{*}Note: `ROM' stands for `Read-Only Memory', and 'MB' stands for 'MegaByte'.}
  \end{center}
\vspace {-1.0em}
\end{table}

\subsection{Real-World Experiment}
We integrated RSMLP into the vehicle-based Retrieval Augmented Generation (RAG) pipeline for real-world experimentation. This system is designed to quickly address issues users encounter while operating their vehicles. In this real-world scenario, referential and elliptical phenomena in multi-turn dialogues lead to difficulties in retrieving the correct documents on the RAG recall side. Our experiments aim to address this issue. As shown in Table \ref{e5}, after integrating RSMLP, the RAG model achieved a recall accuracy of 96\% in multi-turn scenarios, with an inference speed of only 70 milliseconds per sentence. Moreover, our model's ROM usage is only slightly larger than that of BERT, making it highly suitable for edge devices with limited computational power and memory, and demonstrating excellent applicability in such environments.

\section{Conclusions}
In this paper, we propose a simple and efficient model for the IUR task, which utilizes an MLP architecture based on a down-sampling strategy. Our model achieves advanced performance and inference speed on public IUR datasets. Future work will involve exploring the extension of this framework to other conversational domains.

\newpage
\bibliographystyle{IEEEtran}
\bibliography{IEEEabrv,example_paper}

\begin{thebibliography}{10}
\providecommand{\url}[1]{#1}
\csname url@samestyle\endcsname
\providecommand{\newblock}{\relax}
\providecommand{\bibinfo}[2]{#2}
\providecommand{\BIBentrySTDinterwordspacing}{\spaceskip=0pt\relax}
\providecommand{\BIBentryALTinterwordstretchfactor}{4}
\providecommand{\BIBentryALTinterwordspacing}{\spaceskip=\fontdimen2\font plus
\BIBentryALTinterwordstretchfactor\fontdimen3\font minus
  \fontdimen4\font\relax}
\providecommand{\BIBforeignlanguage}[2]{{%
\expandafter\ifx\csname l@#1\endcsname\relax
\typeout{** WARNING: IEEEtran.bst: No hyphenation pattern has been}%
\typeout{** loaded for the language `#1'. Using the pattern for}%
\typeout{** the default language instead.}%
\else
\language=\csname l@#1\endcsname
\fi
#2}}
\providecommand{\BIBdecl}{\relax}
\BIBdecl

\bibitem{b31}
S.~Zhang, ``Personalizing dialogue agents: I have a dog, do you have pets
  too,'' \emph{arXiv preprint arXiv:1801.07243}, 2018.

\bibitem{b32}
H.~Zhou, C.~Zheng, K.~Huang, M.~Huang, and X.~Zhu, ``Kdconv: A chinese
  multi-domain dialogue dataset towards multi-turn knowledge-driven
  conversation,'' \emph{arXiv preprint arXiv:2004.04100}, 2020.

\bibitem{b33}
K.~Xu, H.~Wu, L.~Song, H.~Zhang, L.~Song, and D.~Yu, ``Conversational semantic
  role labeling,'' \emph{IEEE/ACM Transactions on Audio, Speech, and Language
  Processing}, vol.~29, pp. 2465--2475, 2021.

\bibitem{b34}
D.~Yu, K.~Sun, C.~Cardie, and D.~Yu, ``Dialogue-based relation extraction,''
  \emph{arXiv preprint arXiv:2004.08056}, 2020.

\bibitem{b1}
H.~Su, X.~Shen, R.~Zhang, F.~Sun, P.~Hu, C.~Niu, and J.~Zhou, ``Improving
  multi-turn dialogue modelling with utterance rewriter,'' \emph{arXiv preprint
  arXiv:1906.07004}, 2019.

\bibitem{b3}
V.~Kumar and S.~Joshi, ``Non-sentential question resolution using sequence to
  sequence learning,'' in \emph{Proceedings of COLING 2016, the 26th
  International Conference on Computational Linguistics: Technical Papers},
  2016, pp. 2022--2031.

\bibitem{b4}
Z.~Pan, K.~Bai, Y.~Wang, L.~Zhou, and X.~Liu, ``Improving open-domain dialogue
  systems via multi-turn incomplete utterance restoration,'' in
  \emph{Proceedings of the 2019 Conference on Empirical Methods in Natural
  Language Processing and the 9th International Joint Conference on Natural
  Language Processing (EMNLP-IJCNLP)}, 2019, pp. 1824--1833.

\bibitem{b5}
K.~Zhou, K.~Zhang, Y.~Wu, S.~Liu, and J.~Yu, ``Unsupervised context rewriting
  for open domain conversation,'' \emph{arXiv preprint arXiv:1910.08282}, 2019.

\bibitem{b6}
Q.~Liu, B.~Chen, J.-G. Lou, B.~Zhou, and D.~Zhang, ``Incomplete utterance
  rewriting as semantic segmentation,'' \emph{arXiv preprint arXiv:2009.13166},
  2020.

\bibitem{b7}
Y.~Zhang, Z.~Li, J.~Wang, N.~Cheng, and J.~Xiao, ``Self-attention for
  incomplete utterance rewriting,'' in \emph{ICASSP 2022-2022 IEEE
  International Conference on Acoustics, Speech and Signal Processing
  (ICASSP)}.\hskip 1em plus 0.5em minus 0.4em\relax IEEE, 2022, pp. 8047--8051.

\bibitem{b10}
K.~Xu, H.~Tan, L.~Song, H.~Wu, H.~Zhang, L.~Song, and D.~Yu, ``Semantic role
  labeling guided multi-turn dialogue rewriter,'' \emph{arXiv preprint
  arXiv:2010.01417}, 2020.

\bibitem{b11}
J.~Hao, L.~Song, L.~Wang, K.~Xu, Z.~Tu, and D.~Yu, ``Rast: Domain-robust
  dialogue rewriting as sequence tagging,'' in \emph{Proceedings of the 2021
  Conference on Empirical Methods in Natural Language Processing}, 2021, pp.
  4913--4924.

\bibitem{b12}
J.~D. M.-W.~C. Kenton and L.~K. Toutanova, ``Bert: Pre-training of deep
  bidirectional transformers for language understanding,'' in \emph{Proceedings
  of naacL-HLT}, vol.~1, 2019, p.~2.

\bibitem{b14}
D.~Hendrycks and K.~Gimpel, ``Gaussian error linear units (gelus),''
  \emph{arXiv preprint arXiv:1606.08415}, 2016.

\bibitem{b29}
X.~Zeng, D.~Zeng, S.~He, K.~Liu, and J.~Zhao, ``Extracting relational facts by
  an end-to-end neural model with copy mechanism,'' in \emph{Proceedings of the
  56th Annual Meeting of the Association for Computational Linguistics (Volume
  1: Long Papers)}, 2018, pp. 506--514.

\bibitem{b30}
J.~Gu, Z.~Lu, H.~Li, and V.~O. Li, ``Incorporating copying mechanism in
  sequence-to-sequence learning,'' \emph{arXiv preprint arXiv:1603.06393},
  2016.

\bibitem{b17}
S.~Ioffe, ``Batch normalization: Accelerating deep network training by reducing
  internal covariate shift,'' \emph{arXiv preprint arXiv:1502.03167}, 2015.

\bibitem{b18}
A.~Elgohary, D.~Peskov, and J.~Boyd-Graber, ``Can you unpack that? learning to
  rewrite questions-in-context,'' \emph{Can You Unpack That? Learning to
  Rewrite Questions-in-Context}, 2019.

\bibitem{b25}
A.~See, P.~J. Liu, and C.~D. Manning, ``Get to the point: Summarization with
  pointer-generator networks,'' \emph{arXiv preprint arXiv:1704.04368}, 2017.

\bibitem{b26}
D.~Bahdanau, ``Neural machine translation by jointly learning to align and
  translate,'' \emph{arXiv preprint arXiv:1409.0473}, 2014.

\bibitem{b9}
K.~Xu, H.~Tan, L.~Song, H.~Wu, H.~Zhang, L.~Song, and D.~Yu, ``Semantic role
  labeling guided multi-turn dialogue rewriter,'' \emph{arXiv preprint
  arXiv:2010.01417}, 2020.

\bibitem{b20}
M.~Huang, F.~Li, W.~Zou, and W.~Zhang, ``Sarg: A novel semi autoregressive
  generator for multi-turn incomplete utterance restoration,'' in
  \emph{Proceedings of the AAAI Conference on Artificial Intelligence},
  vol.~35, no.~14, 2021, pp. 13\,055--13\,063.

\bibitem{b21}
K.~Papineni, S.~Roukos, T.~Ward, and W.-J. Zhu, ``Bleu: a method for automatic
  evaluation of machine translation,'' in \emph{Proceedings of the 40th annual
  meeting of the Association for Computational Linguistics}, 2002, pp.
  311--318.

\bibitem{b22}
C.-Y. Lin, ``Rouge: A package for automatic evaluation of summaries,'' in
  \emph{Text summarization branches out}, 2004, pp. 74--81.

\bibitem{b23}
T.~Wolf, L.~Debut, V.~Sanh, J.~Chaumond, C.~Delangue, A.~Moi, P.~Cistac,
  T.~Rault, R.~Louf, M.~Funtowicz \emph{et~al.}, ``Transformers:
  State-of-the-art natural language processing,'' in \emph{Proceedings of the
  2020 conference on empirical methods in natural language processing: system
  demonstrations}, 2020, pp. 38--45.

\bibitem{b24}
P.~K. Diederik, ``Adam: A method for stochastic optimization,'' \emph{(No
  Title)}, 2014.

\end{thebibliography}

\end{CJK}
\end{document}